\documentclass[10pt,twocolumn,letterpaper]{article}
\usepackage{bbm}
\usepackage[accsupp]{axessibility} 
\usepackage{graphicx}
\usepackage{amsmath}
\usepackage{amssymb}
\usepackage{amsthm}
\usepackage{mathrsfs}
\usepackage{amsfonts}
\usepackage{booktabs}
\usepackage{multirow}
\usepackage{multicol}
\usepackage{ftnxtra}
\usepackage{enumitem}
\usepackage{fnpos} % \makeFNbelow by default 
\usepackage[table]{xcolor}
\usepackage{tablefootnote}
\usepackage{color, colortbl}
\usepackage[section]{placeins}
\usepackage{float}
\usepackage{lipsum}
\usepackage{pifont}% http://ctan.org/pkg/pifont
\newcommand{\cmark}{\ding{51}}%
\newcommand{\xmark}{\ding{55}}%
\usepackage[pagenumbers]{cvpr} % To force page numbers, e.g. for an arXiv version
\def\vs{\emph{vs.}}
\def\etc{\emph{etc.}}
\definecolor{cvprblue}{rgb}{0.21,0.49,0.74}
\usepackage[pagebackref,breaklinks,colorlinks,citecolor=cvprblue]{hyperref}
\hypersetup{citecolor=[RGB]{119,185,0}}
\usepackage[capitalize]{cleveref}
\def\paperID{3514} % *** Enter the Paper ID here
\def\confName{CVPR}
\def\confYear{2024}

\title{Is Ego Status All You Need for Open-Loop End-to-End Autonomous Driving? }
\author{
\text{Zhiqi Li}$^{1,2}$\thanks{Work done during an internship at NVIDIA.}~,
\text{Zhiding Yu}$^{2}$\thanks{Corresponding author: \href{mailto:zhidingy@nvidia.com}{zhidingy@nvidia.com}.}~,
\text{Shiyi Lan}$^{2}$,
\text{Jiahan Li}$^{1}$,
\text{Jan Kautz}$^{2}$,
\text{Tong Lu}$^{1}$,
\text{Jose M. Alvarez}$^{2}$
\\ [0.15cm]
$^1$National Key Lab for Novel Software Technology, Nanjing University~~~~$^2$NVIDIA
}

\begin{document}
\maketitle
\begin{abstract}
End-to-end autonomous driving recently emerged as a promising research direction to target autonomy from a full-stack perspective. Along this line, many of the latest works follow an open-loop evaluation setting on nuScenes to study the planning behavior. In this paper, we delve deeper into the problem by conducting thorough analyses and demystifying more devils in the details. We initially observed that the nuScenes dataset, characterized by relatively simple driving scenarios, leads to an under-utilization of perception information in end-to-end models incorporating ego status, such as the ego vehicle's velocity. These models tend to rely predominantly on the ego vehicle's status for future path planning. 
Beyond the limitations of the dataset, we also note that current metrics do not comprehensively assess the planning quality, leading to potentially biased conclusions drawn from existing benchmarks. To address this issue, we introduce a new metric to evaluate whether the predicted trajectories adhere to the road. 
We further propose a simple baseline able to achieve competitive results without relying on perception annotations.
Given the current limitations on the benchmark and metrics, we suggest the community reassess relevant prevailing research and be cautious about whether the continued pursuit of state-of-the-art would yield convincing and universal conclusions. Code and models are available at  \url{https://github.com/NVlabs/BEV-Planner}.
\end{abstract}    
\section{Introduction}
\label{sec:intro}

\begin{figure}[t]
\centering
\includegraphics[width=1\columnwidth]{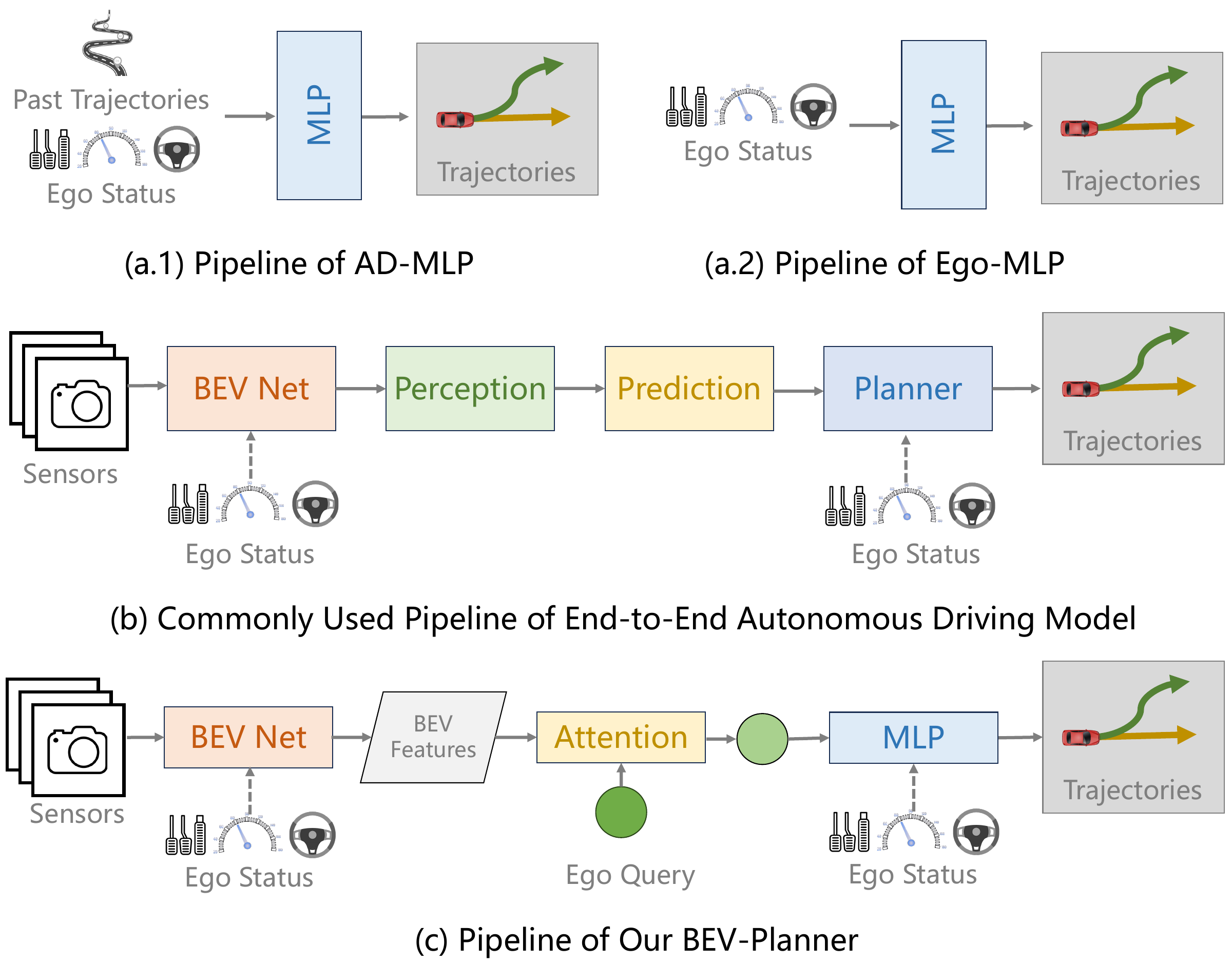}\\
\vspace{-5pt}
\caption{
(a) AD-MLP uses both ego status and past trajectory GTs as input. Our reproduced version (Ego-MLP) drops the past trajectories. (b) The existing end-to-end autonomous driving pipeline consists of perception, prediction, and planning modules. Ego status can be integrated into the bird's-eye view (BEV) generation module or within the planning module. (c) We design a simple baseline for comparison with existing methods. The simple baseline does not leverage the perception or prediction module and directly predicts the final trajectories based on BEV features.
}
\label{fig:e2e_pipeline}
\end{figure}

End-to-end autonomous driving aims to jointly consider perception and planning in a full-stack manner~\cite{chen2023end, prakash2021multi, 
bojarski2016end, tampuu2020survey}. An underlying motivation is to evaluate autonomous vehicle (AV) perception as a means to an end (planning), instead of overfitting to certain perception metrics.

Unlike perception, the planning is generally much more open-ended 
and hard to quantify~\cite{claussmann2019review, dolgov2008practical}.
This open-ended nature of planning would ideally favor a closed-loop evaluation setting where other agents could react to the behavior of the ego vehicle, and the raw sensor data could also change accordingly. However, both agent behavior modeling and real-world data simulation within closed-loop simulators~\cite{dosovitskiy2017carla, li2022metadrive} remain challenging open problems to date. 
As such, closed-loop evaluation 
inevitably introduces considerable domain gaps to the real world.

Open-loop evaluation, on the other hand, aims to treat human driving as the ground truth and formulate planning as imitation learning~\cite{hu2023uniad}. Such formulation allows the readily usage of real-world datasets via simple log-replay, avoiding the domain gaps from simulation. It also offers other advantages, such as the capacity to train and validate models in complex and diverse traffic scenarios, which are often difficult to generate with high fidelity in simulations~\cite{chen2023end}. For these benefits, a well-established body of research focuses on open-loop end-to-end autonomous driving with real-world dataset~\cite{hu2022stp3, hu2023uniad, jiang2023vad, ye2023fusionad, caesar2020nuscenes}.

Current prevailing end-to-end autonomous driving methods~\cite{hu2023uniad, hu2022stp3, jiang2023vad, ye2023fusionad} commonly use nuScenes~\cite{caesar2020nuscenes} for open-loop evaluation of their planning behavior. For instance, UniAD~\cite{hu2023uniad} studies the influence of different perception task modules to the final planning behavior. However, AD-MLP~\cite{zhai2023admlp} recently points out that a simple MLP network can also achieve state-of-the-art planning results, relying solely on the ego status information. This motivates us to ask an important question: 

\begin{itemize}
    \vspace{2mm}
    \item[] \textit{Is ego status all you need for open-loop end-to-end autonomous driving?}
    \vspace{2mm}
\end{itemize}

\noindent Our answer is \textbf{yes and no}, considering both the pros and cons of using ego status in current benchmarks:

\vspace{2mm}
\noindent\textbf{Yes.} Information such as velocity, acceleration and yaw angle in the ego status should apparently benefit the planning task. To verify this, we fix an open issue\footnote{\url{https://github.com/E2E-AD/AD-MLP/issues/4}.} of AD-MLP and remove the use of history trajectory ground truths (GTs) to prevent potential label leakage. Our reproduced model, Ego-MLP (\cref{fig:e2e_pipeline} a.2), relies solely on the ego status and is on par with state-of-the-art methods in terms of existing L2 distance and collision rate metrics. Another observation is that only existing methods~\cite{jiang2023vad,hu2023uniad,ye2023fusionad}, which incorporate ego status information within the planner module, can obtain results on par with Ego-MLP. Although these methods employ additional perception information (tracking, HD map, \etc), they don't demonstrate superiority compared to Ego-MLP. These observations verify the dominating role of ego status in the open-loop evaluation of end-to-end autonomous driving.

\vspace{2mm}
\noindent\textbf{And No.}
It is also evident that autonomous driving as a safety-critical application should not depend solely on ego status for decision-making. So why does this phenomenon occur where using only ego status can achieve state-of-the-art planning results? To address the question, we present a comprehensive set of analyses covering existing open-loop end-to-end autonomous driving methods.  We identify major shortcomings within existing research, including aspects related to datasets, evaluation metrics, and specific model implementations. We itemize and detail these shortcomings in the rest of the section:

\begin{figure}[t]
\centering
\includegraphics[width=1\columnwidth]{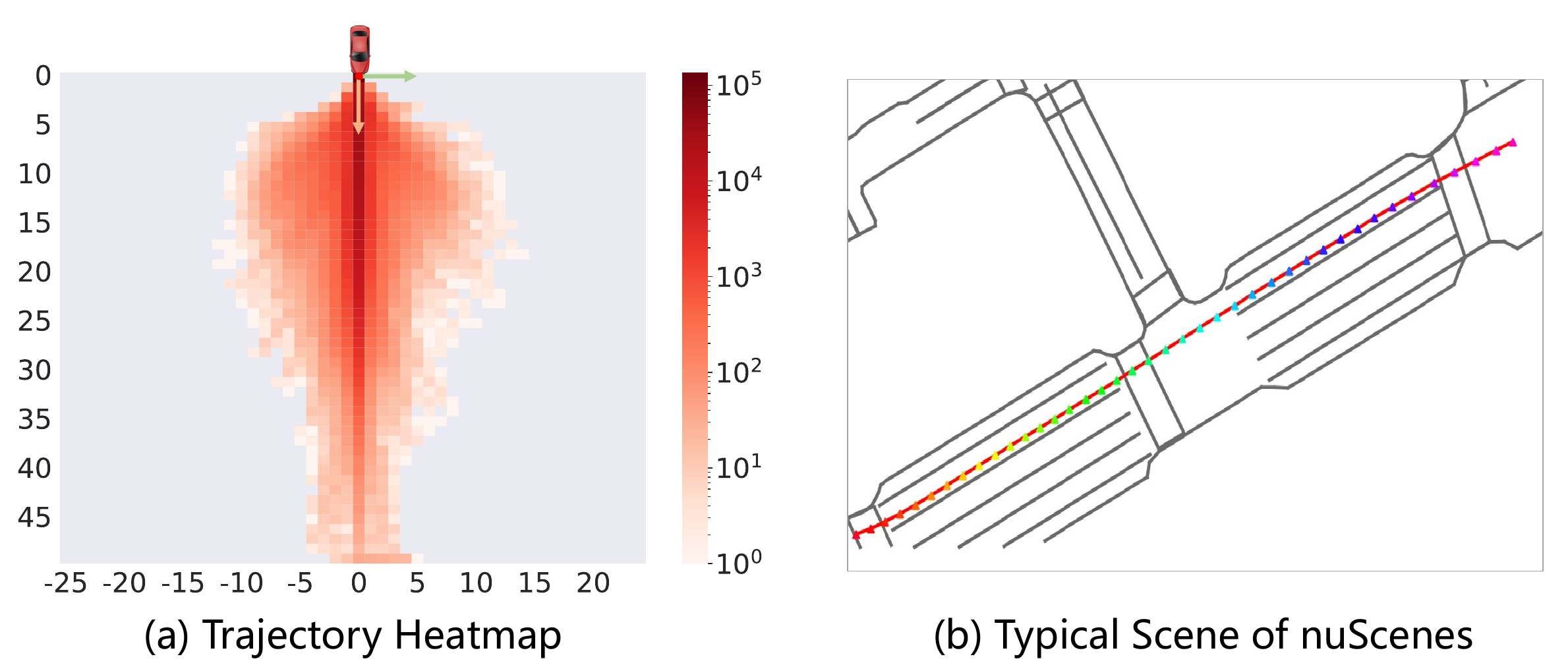}
\caption{(a) The ego car trajectory heatmap on nuScenes dataset. (b) The majority of the scenes within the nuScenes dataset consist of straightforward driving situations.}
\vspace{-10pt}
\label{fig:traj_heatmap}
\end{figure}

\vspace{2mm}
\noindent\textbf{Imblanced dataset.} 
NuScenes is a commonly used benchmark for open-loop evaluation tasks\cite{hu2021safe,hu2022stp3,jiang2023vad,hu2023uniad,ye2023fusionad,khurana2022differentiable}. However, our
analysis shows that 73.9\% of the nuScenes data involve scenarios of driving straightforwardly, as reflected by the distribution of the trajectory in \cref{fig:traj_heatmap}. For these straight-driving scenarios, maintaining the current velocity, direction, or turning rate can be sufficient most of the time. Hence, ego status information can be easily leveraged as a shortcut to fit the planning task, leading to the strong performance of Ego-MLP on nuScenes.

\noindent\textbf{Existing metrics are not comprehensive.}
The remaining 26.1\% of nuScenes data involve more challenging driving scenarios for potentially better benchmarks of planning behaviors. However, we argue that the widely used current metrics, such as the L2 distance between prediction and planning GT and the collision rates between ego vehicle and surrounding obstacles, fail to accurately measure the model's planning behavior quality.
Through visualizing numerous predicted trajectories generated from various methods, we note that some highly risky trajectories, such as running off the road may not get severely penalized in existing metrics.
In response to this issue, we introduce a new metric to calculate the interaction rate between the predicted trajectories and the road boundaries. 
While focusing on intersection rates with road boundaries, the benchmark will experience a substantial transformation. In terms of this new metric, Ego-MLP tends to predict trajectories that deviate from the road more frequently than UniAD. 

\noindent\textbf{Ego status bias against driving logic.} With ego status being a potential source causing overfitting, we further observe an interesting phenomenon. Our experiment results suggest that, in some cases, completely removing visual input from an existing end-to-end autonomous driving framework does not significantly degrade the planning behavior. This contradicts the basic driving logic in the sense that perception is expected to provide useful information for planning. For instance, blanking all the camera input in VAD~\cite{jiang2023vad} leads to complete failure of the perception module but minor degrade in planning, when ego status is present. However, altering the input ego velocity can significantly influence the final predicted trajectory.

In conclusion, we conjecture that recent efforts in end-to-end autonomous driving and their state-of-the-art scores on nuScenes are likely to be caused by the over-reliance on ego status, coupled with the dominance of simple driving scenarios. Furthermore, current metrics fall short in comprehensively assessing the quality of model-predicted trajectories. These open issues and shortcomings may have under-presented the potential complexity of the planning task and created a \textbf{misleading impression} that ego status is all you need for open-loop end-to-end autonomous driving.

The potential interference of ego status in current open-loop end-to-end autonomous driving research raises another question: Is it possible to negate this influence by removing ego status from the whole model? However, it's important to note that even excluding the impact of ego status, \textit{the reliability of open-loop autonomous driving research based on the nuScenes dataset remains in question.}

\section{Related Work}
\label{sec:formatting}

%-------------------------------------------------------------------------
\subsection{BEV perception}
In recent years, BEV-based autonomous driving perception methods have made great progress. Lift-Splat-Shoot~\cite{philion2020lift} firstly propose to use latent depth distribution to perform view transformation. BEVFormer~\cite{li2022bevformer} introduces temporal clues into BEV perception and greatly boosts the 3D detection performance. A series of subsequent works~\cite{huang2022bevdet4d, li2022bevdepth, 
huang2023leveraging, park2022time, liu2022petrv2, li2023fb, xie2022m, lin2023sparse4d, liu2023sparsebev, li2023fbocc, yang2023bevformer} obtain more accurate 3D perception results by obtaining more accurate depth information or making better use of temporal information. The incorporation of temporal information typically necessitates the alignment of features across different timesteps~\cite{li2022bevformer, li2022delving, huang2022bevdet4d, wang2023exploring}. In the alignment process, the ego status is either implicitly encoded within the input feature~\cite{wang2023exploring} or is explicitly used to translate BEV features~\cite{huang2022bevdet4d}. Methods~\cite{liao2022maptr,chen2022persformer, wang2023bev, li2021hdmapnet, liu2023vectormapnet, yuan2023streammapnet} explored map perception based on BEV features.

%-------------------------------------------------------------------------
\subsection{End-to-end autonomous driving}
Modern autonomous driving systems are usually divided into three main tasks: perception, prediction, and planning. End-to-end autonomous driving that directs learning from raw sensor data to planning trajectories or driving commands eliminates the need for manual feature extraction, leading to efficient data utilization and adaptability to diverse driving scenarios. There exists a body of research~\cite{wang2022interfusion, shao2023interfuser, wu2022trajectory} focused on closed-loop end-to-end driving within simulators~\cite{dosovitskiy2017carla, li2022metadrive}. However, a domain gap persists between the simulator environment and the real world, particularly concerning sensor data and the motion status of agents. Recently, open-loop end-to-end autonomous driving has attracted more attention.
End-to-end autonomous driving methods~\cite{jiang2023vad, hu2023uniad, ye2023fusionad, sadat2020perceive, casas2021mp3, gu2023vip3d}  that involve learning intermediate tasks claim their effectiveness in improving final planning performance. AD-MLP~\cite{zhai2023admlp} pointed out the issue of imbalanced data distribution in nuScenes and attempted to use only ego status as the model input to achieve arts performance. However, AD-MLP benefits from utilizing the historical trajectory of the ego car as input. Given that none of the existing methods use the historical trajectory information of the ego car, we argue that using the historical trajectory in open-loop autonomous driving is a subject of debate, as the model itself does not generate this historical trajectory but rather by an actual human driver.

\section{Proposed BEV-Planner}
In fact, ST-P3~\cite{hu2022stp3}, a previous method that often serves as a baseline, uses partially incorrect GT data during training and evaluation\footnote{https://github.com/OpenDriveLab/ST-P3/issues/24}. Consequently, when conducting comparisons between other methods and ST-P3, the validity of the conclusions drawn must be carefully evaluated. Therefore, in this paper, it is necessary for us to redesign a baseline method to compare with existing methods.
At the same time, to better explore the impact of ego status, we also need a relatively clear baseline method. Based on these considerations, we have designed a very simple baseline in this paper, named \textbf{BEV-Planner}, as shown in the ~\cref{fig:e2e_pipeline}(c). For our pipeline, we first generate the BEV feature and concatenate it with history BEV features, mainly following the previous method~\cite{hu2022stp3,huang2022bevdet4d, li2022bevdepth}. Please note while concatenating BEV features from different timesteps, we didn't perform feature alignment.
After obtaining the BEV features, we directly perform a cross-attention~\cite{vaswani2017attention} between the BEV features and the ego query, which is a learnable embedding. The final trajectories are predicted based on the refined ego query through MLPs. The process can be formulated as follows:
\begin{equation}
    \tau\!=\!\textbf{MLP}(\textbf{attn}(q\!=\!Q, k\!=\!B, v\!=\!B)),
\end{equation}
where $Q$ is the ego query, $B$ is the BEV features after temporal fusion. $\tau$ is the final predicted trajectories.

To align with existing methods, we also designed baseline approaches that incorporate ego status into the BEV or planner modules. The strategy of incorporating ego status into the BEV aligns with previous approaches~\cite{li2022bevformer,hu2023uniad, jiang2023vad}. The strategy of incorporating ego status in the planner is directly concatenating the ego query with a vector containing ego status.

Compared to existing methods, this simple method didn't require any human-labeled data, including bounding boxes, tracking IDs, HD maps, \etc ~ For this proposed baselines, we only use one L1 loss for trajectory supervision. We wish to underscore that our proposed baseline method is not intended for real-world deployment, owing to its deficiencies in providing adequate constraints and interoperability.

\section{Experiments}

\subsection{Implementation Details}

Our baseline uses an R50 backbone~\cite{he2016deep}. The input resolution is $256\!\times\! 704$, smaller than existing methods~\cite{jiang2023vad, hu2023uniad}. The BEV resolution is $128\!\times\!128$ with a perception range of around 50 meters. For the baseline that uses history BEV features, we directly concatenate the BEV features from the past 4 timesteps to current BEV features along the channel dimension without alignment. A BEV encoder from method~\cite{huang2022bevdet4d} is further used to squeeze the channel dimension to 256. We train our model for 12 epochs on 8 V100 GPUs, with a batch size of 32 and a learning rate of 1e-4.

\subsection{Metrics}

In the Appendix, we will introduce the shortcomings of the currently commonly used  collision rate and another metric used to evaluate the smoothness of predicted trajectory.

\paragraph{Curb Collision Rate (CCR).}
In this study, to more comprehensively assess the quality of predicted trajectories, we employed a new metric that calculates the collision rate between the predicted trajectories and curbs (road boundaries). Staying on the road is vital for the safety of autonomous driving systems, yet existing evaluation metrics overlook the integration of map priors. Intuitively, safe trajectories should avoid collision with curbs. Our Curb Collision Rate (CCR) typically indicate the possibility of leaving the drivable area, which can pose safety hazards. We recognize that certain annotated road boundaries on nuScenes are indeed traversable, and ground truth trajectories may intersect with these boundaries under specific conditions. However, from a statistical viewpoint, this metric can effectively represent the overall rationality of the model's predicted trajectories.  The implementation of the CCR metric is informed by the collision rate. To facilitate this, we rasterize the road boundary using a resolution of 0.1 meters. More details are in the Appendix.

\paragraph{Union Implementation.} 
Given the lack of a standardized approach to assessment metrics, methodologies might differ in the nuances of their metric executions. In this study, we utilized the official open-source repositories from various methodologies to produce predicted trajectories. Subsequently, we employed a consistent metric implementation across all methods for evaluation, guaranteeing equity. Addressing concerns raised by AD-MLP~\cite{zhai2023admlp} regarding the potential for false collisions due to a coarse-grained grid size (0.5m), we adopt a finer default grid size of 0.1m in our work to mitigate this issue.

\begin{table*}
\begin{center}
\resizebox{\textwidth}{!}{
\setlength{\tabcolsep}{6pt}
\begin{tabular}{c|l|cc|cccc|cccc|cccc|c}
\toprule
\multirow{2}{*}{ID}&
\multirow{2}{*}{Method} &
\multicolumn{2}{c|}{Ego Status} &
\multicolumn{4}{c|}{L2 (m) $\downarrow$} & 
\multicolumn{4}{c|}{Collision (\%) $\downarrow$} &
\multicolumn{4}{c|}{CCR (\%) $\downarrow$} &
\multirow{2}{*}{ckpt. source}  \\
 &&in BEV&in Planer& 1s & 2s & 3s &\cellcolor{gray!30}Avg. & 1s & 2s & 3s& 
\cellcolor{gray!30}Avg. & 1s & 2s & 3s &\cellcolor{gray!30}Avg.\\
\midrule
0&ST-P3 &\xmark &\xmark &\textcolor{gray}{1.59}$^{\text{\textdagger}}$&\textcolor{gray}{2.64}$^{\text{\textdagger}}$&\textcolor{gray}{3.73}$^{\text{\textdagger}}$&\textcolor{gray}
{2.65}$^{\text{\textdagger}}$&\textcolor{gray}{0.69}$^{\text{\textdagger}}$&\textcolor{gray}{3.62}$^{\text{\textdagger}}$&\textcolor{gray}
{8.39}$^{\text{\textdagger}}$& \textcolor{gray}
{4.23}$^{\text{\textdagger}}$& \textcolor{gray}
{2.53}$^{\text{\textdagger}}$&\textcolor{gray}{8.17}$^{\text{\textdagger}}$&\textcolor{gray}
{14.4}$^{\text{\textdagger}}$&\textcolor{gray}{8.37}$^{\text{\textdagger}}$&Official \\
1\label{id1}&UniAD &\xmark& \xmark& 0.59 & 1.01 & 1.48 & 1.03& 0.16 & 0.51 & 1.64 &0.77 &0.35&1.46&3.99& 1.93&Reproduce  \\
2&UniAD &\cmark& \xmark& 0.35 & 0.63 & 0.99&0.66  & 0.16 & 0.43 & 1.27&0.62 &0.21&\textbf{1.32}&3.63& 1.72 & Official  \\
3&UniAD &\cmark& \cmark& 0.20 & 0.42 & 0.75& 0.46 & 0.02 & \textbf{0.25} & 0.84&0.37 &\textbf{0.20}&1.33&\textbf{3.24}& \textbf{1.59}& Reproduce  \\
4&VAD-Base &\xmark&\xmark& 0.69 & 1.22 & 1.83 &1.25 & 0.06 & 0.68 & 2.52 &1.09 &1.02&3.44&7.00& 3.82&Reproduce  \\
5&VAD-Base&\cmark&\xmark& 0.41 & 0.70 & 1.06&0.72 & 0.04 & 0.43 & 1.15 &0.54&0.60&2.38&5.18&2.72 & Official   \\
6&VAD-Base &\cmark&\cmark& 0.17 & 0.34 & 0.60& 0.37 & 0.04 & {0.27} & \textbf{0.67}&\textbf{0.33} &0.21 &2.13&5.06& 2.47&Official   \\
\midrule
7&GoStright & - &\cmark& 0.38& 0.79& 1.33&0.83  & 0.15 & 0.60 & 2.50&1.08 &2.07&8.09&15.7& 8.62& - \\ 
8&Ego-MLP& -& \cmark& \textbf{0.15} & \textbf{0.32} & 0.59  & \textbf{0.35}&\textbf{0.00} & {0.27} & 0.85&0.37 &0.27&2.52&6.60& 2.93 \\
\midrule

9& BEV-Planner* &\xmark &\xmark & 0.27 & 0.54&0.90 & 0.57& 0.04 & 0.35 & 1.80 &0.73&0.63&3.38&7.93 &3.98 &- \\
% \midrule
10& BEV-Planner &\xmark &\xmark & 0.30 & 0.52&0.83 &0.55 & 0.10 & 0.37 & 1.30 &0.59&0.78&3.79& 8.22&4.26&-\\
11& \textcolor{gray}{BEV-Planner+} &\cmark &\xmark & 0.28 & 0.42&0.68 & 0.46& 0.04 & 0.37&  1.07&0.49 & 0.70&3.77&8.15&4.21&-\\
12& \textcolor{gray}{BEV-Planner++} &\cmark &\cmark & 0.16 & 0.32&\textbf{0.57} & \textbf{0.35}& \textbf{0.00} & 0.29 & 0.73 &0.34 &0.35&2.62&6.51&3.16 &- \\
% \midrule
\bottomrule
\end{tabular}}
\end{center}
\vspace{-10pt}
\caption{
\textbf{Open-loop planning performance. }\text{\textdagger}: The official implementation of ST-P3 (ID-0) utilized partial erroneous ground truth trajectories, with details provided in the appendix. The official UniAD (ID-2) utilized ego status in its BEV module. It is of particular note that the performance of the officially open-sourced model exceeds the results reported in the original paper~\cite{hu2023uniad}. 
We implemented minor modifications to the official codebases of UniAD and VAD to investigate the variations in results arising from different applications of ego status (ID-1, 3 \& 4). A naive strategy (ID-7) of proceeding at the current speed also yields satisfactory results. Without the perception module, Ego-MLP (ID-8), utilizing solely ego velocity, acceleration, yaw angle, and driving command, achieves performance on par with current state-of-the-art models on previous L2 distance and collision rate metrics. *: Our simple baseline (ID-9) didn't utilize historical temporal information. The baseline (ID-10) utilizes the temporal clues from the past 4 frames. To ensure the comprehensiveness of our experiment, we also conduct investigations into the influence of ego status on our baseline model (ID-11\& 12).
}
\label{tab:sota-plan}
 \end{table*}

\begin{table*}[]
\begin{center}
\resizebox{\textwidth}{!}{
\setlength{\tabcolsep}{5pt}
\begin{tabular}{l|c|c| cccc|cccc|cccc|cc}
\toprule
\multirow{2}{*}{Method} &
{Img } &
Ego Status &
\multicolumn{4}{c|}{L2 (m) $\downarrow$} & 
\multicolumn{4}{c|}{Collision (\%) $\downarrow$} &
\multicolumn{4}{c|}{CCR (\%) $\downarrow$} &
Det. & Map\\
 &Corruption&Noise& 1s & 2s & 3s & \cellcolor{gray!30}Avg. & 1s & 2s & 3s & \cellcolor{gray!30}Avg.& 1s & 2s & 3s & \cellcolor{gray!30}Avg. &(NDS) &(mAP) \\
\midrule
VAD-Base*& - & - & 0.41 & 0.70 & 1.06 & 0.72 & 0.04 & 0.43 & 1.15& 0.54& 0.60&2.38&5.18&2.72&\textbf{46.0} & \textbf{47.5}\\
VAD-Base & - & -& \textbf{0.17} & \textbf{0.34} & \textbf{0.60} & \textbf{0.37} & 0.04 & {0.27} & {0.67}&0.33&0.21&2.13&5.06 & 2.47& 45.5 &  47.0\\
\midrule
VAD-Base & Snow & -& 0.19 & 0.41 & 0.76 & 0.45 & \textbf{0.00} & {0.20} & {0.76}& 0.32&0.21 &2.27&5.98&2.82&36.1 &29.4  \\
VAD-Base & Fog & -& 0.19 & 0.40 & 0.75 & 0.45 & 0.02 & {0.20} & {0.68}& 0.30& 0.23&2.21&5.90&2.78&34.3 & 29.4 \\
VAD-Base &  Glare& -& 0.19 & 0.40 & 0.74 & 0.44 & 0.02 & \textbf{0.18} & \textbf{0.59}& \textbf{0.26}& 0.21&2.11&5.57&2.63& 41.7 & 38.3 \\
VAD-Base &  Rain & -& 0.19 & 0.41 & 0.75 & 0.45& 0.02 & \textbf{0.18} & {0.66}&0.29 &0.31 &2.38&5.98&2.89& 29.1 &13.0  \\
VAD-Base &  Blank & -& 0.19 & 0.41 & 0.77 & 0.46 & \textbf{0.00} & {0.40} & {1.21}&0.54 &0.35&3.05&7.73&3.71& 0.0 &0.0  \\% 

\midrule
VAD-Base &  - &$v\!\times\! 0.0$ & 3.81 & 6.19 & 8.48 & 6.16 & 1.00 & 6.76 & 16.18& 7.98&\textbf{0.19} & \textbf{0.41}&\textbf{3.10}&\textbf{1.23}&45.5 &47.0  \\% 
VAD-Base &  - &$v\!\times\! 0.5$ & 1.95 & 3.20 & 4.41 & 3.19 & 0.02 & 1.00 & 4.10& 1.71&0.37&2.56&5.57&2,83& 45.5 &47.0  \\% 
VAD-Base &  - &$v\!\times\! 1.5$ & 1.94 & 3.20 & 4.47 & 3.20 & 0.14 & 2.89 & 6.21& 3.08& 1.54&6.29&13.2&7.01&45.5 &47.0  \\% 
VAD-Base &  - &$v\!=\!100 m/s$ & 113 & 206 & 306 & 208 &\textcolor{gray}{8.00}$^{\text{\textdagger}}$& 
\textcolor{gray}{9.66}$^{\text{\textdagger}}$&  \textcolor{gray}{10.49} $^{\text{\textdagger}}$ &\textcolor{gray}{9.38}$^{\text{\textdagger}}$& 
\textcolor{gray}{24.8}$^{\text{\textdagger}}$&\textcolor{gray}{27.5} $^{\text{\textdagger}}$& \textcolor{gray}{28.8}$^{\text{\textdagger}}$&\textcolor{gray}{27.0}$^{\text{\textdagger}}$&45.5 &47.0  \\% 
\bottomrule

\end{tabular}}
\end{center}
\vspace{-10pt}
\caption{\textbf{The VAD-base model's robustness to images and ego status.} To ascertain the impact of perceptual information and ego status on the ultimate planning performance, we systematically introduced noise into each component separately. We utilize the official VAD-Base checkpoint that uses ego status in its planner module. *: the results of VAD-Base without ego status in its planner.
We can observe that introducing corruption to images markedly affects the perception results, especially in the case of using blank images; nonetheless, this does not markedly disrupt the ultimate planning results.
In contrast to the minor impact of image corruption on planning, modifications to the ego vehicle's velocity have a significant effect on the planning results.
Experimental results reveal that in an end-to-end model incorporating both ego status and perceptual information, decision-making is disproportionately influenced by ego status, thereby substantially increasing the model's safety risks. \text{\textdagger}: The collision rate is not precise as the ego car may have departed from the local BEV area. When the input velocity is zero, the model produces almost stationary trajectories, resulting in excellent performance in the CCR. This can be seen as a limitation of the CCR metric.}

\label{tab:noise}
\vspace{-10pt}
\end{table*}

\begin{figure*}[t]
	\centering
	\includegraphics[width=0.9\linewidth]{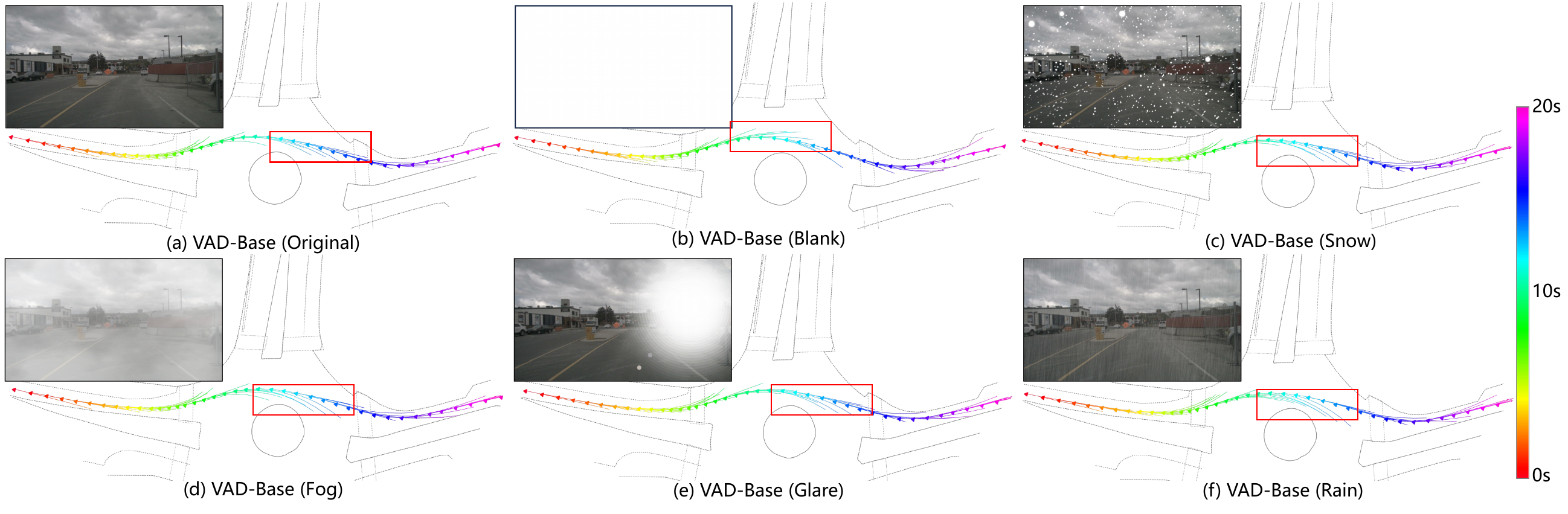}
   % \vskip -0.1in
   \vspace{-5pt}
   \caption {We exhibit the predicted trajectories of the VAD model (incorporating ego status in its planner) under various image corruptions. All trajectories within a given scene (spanning 20 seconds) are presented in the global coordinate system. Each triangular marker signifies a ground truth trajectory point of the ego vehicle, with different colors representing distinct timesteps. Notably, the model's predicted trajectory maintains plausibility, even when blank images serve as input. The trajectories within the red boxes, however, are suboptimal, as further elucidated in Appendix. While corruptions were applied to all surround-view images, for the sake of visualization, only the corresponding front-view images at the initial timestep are displayed.}
  %  \vspace{-0.20in}
\vspace{-10pt}  
\label{fig:vad_noise}
\end{figure*}

\begin{figure}[t]
	\centering
	\includegraphics[width=0.8\columnwidth]{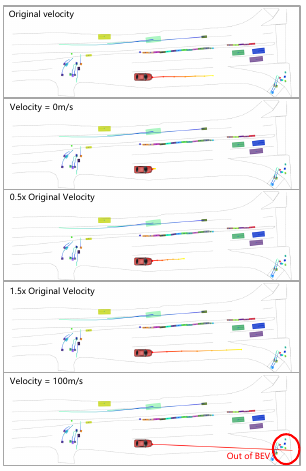}
   % \vskip -0.1in
   %\vspace{-3mm}

   \caption{
For the VAD-based model that incorporates ego status in its planner, we introduce the noise to the ego velocity with the visual inputs remaining constant. Notably, when the velocity data of the ego vehicle is perturbed, the resulting trajectories exhibit substantial alterations. Setting the vehicle's speed to zero results in a stationary prediction, while a speed of 100 m/s leads to the projection of an implausible trajectory. This indicates a disproportionate dependence of the model's decision-making process on the ego status, even though the perception module continues to provide accurate surrounding information.}
\label{fig:velo_noise}
\vspace{-10pt}
\end{figure}

\begin{figure}[htb]
	\centering
	\includegraphics[width=0.9\columnwidth]{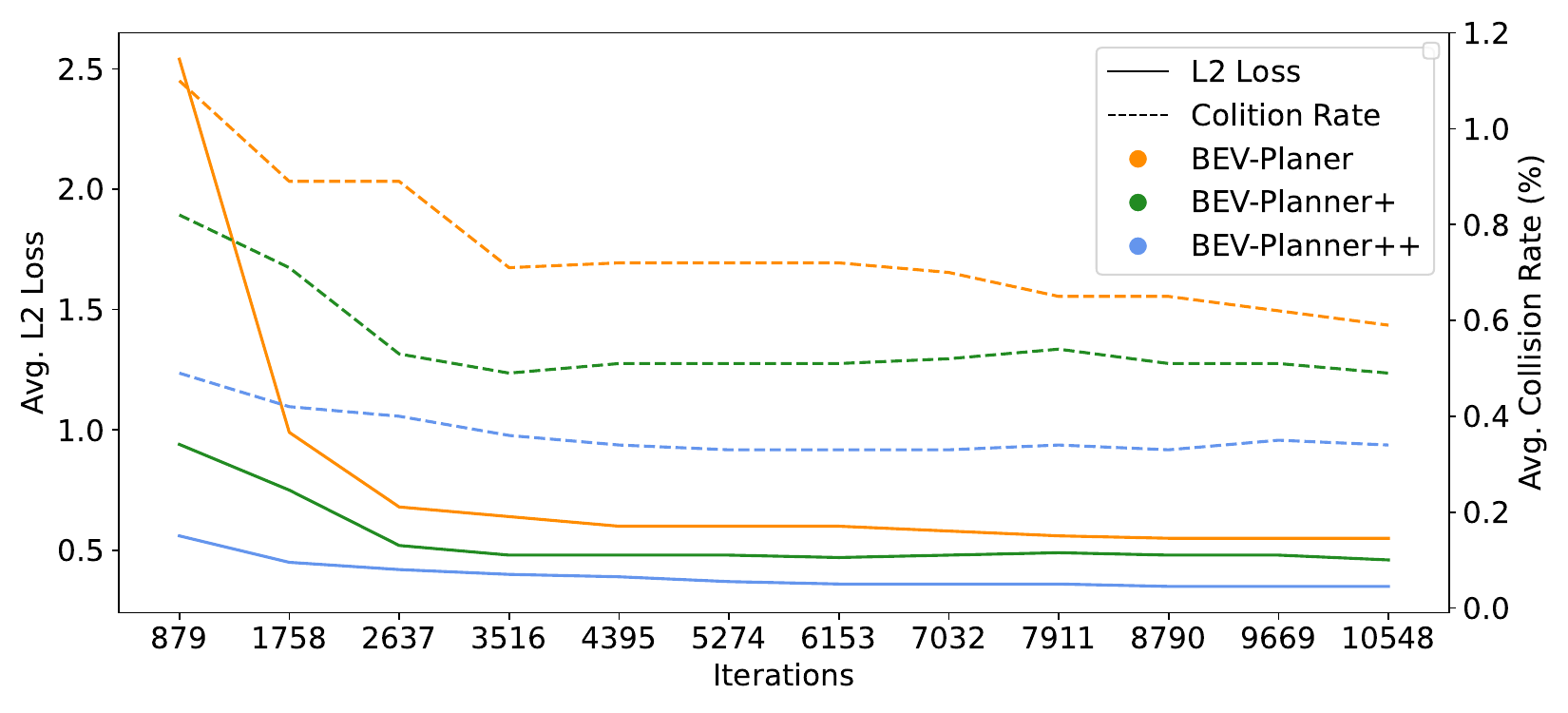}
   % \vskip -0.1in
     \vspace{-10pt}
     \caption{
   Introducing ego status in the BEV-Planner++ enables the model to converge very rapidly. 
}
  \vspace{-10pt}  %  \vspace{-0.20in}
\label{fig:convg}
\end{figure}

\begin{figure}[htb]
	\centering
	\includegraphics[width=0.8\columnwidth]{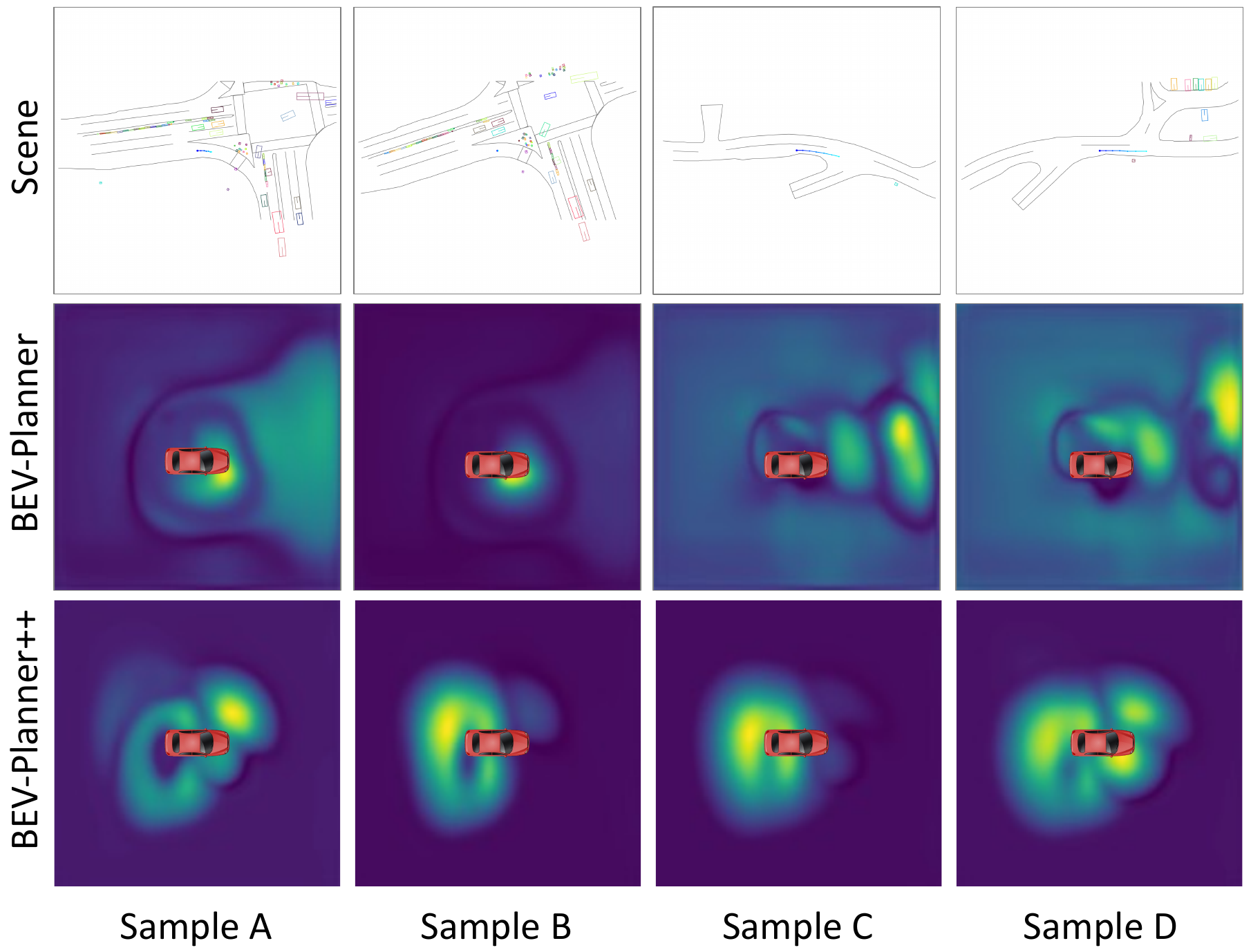}
   % \vskip -0.1in
\textit{}   \caption{Comparing the BEV features of our baselines with the corresponding scenes.}
  %  \vspace{-0.20in}
\label{fig:bev_scenes}
\end{figure}
\vspace{-5pt}
\subsection{Discussion}

\paragraph{Ego status plays a key role.} While focusing solely on previous metrics \textbf{L2 distance} and \textbf{collision rate}, it is observed that the simple strategy (ID-7), which simply continues straight at the current velocity, achieves surprisingly good results. The Ego-MLP model, which didn't leverage perception clues, is actually on par with UniAD and VAD, which use more complex pipelines.  From another perspective, it is observed that existing methods can only match the performance of Ego-MLP when the ego vehicle's status is incorporated into the planner. In contrast, reliance solely on camera inputs leads to results significantly inferior to those achieved by Ego-MLP. Considering these observations, we may tentatively infer an intriguing conclusion: utilizing a combination of sensory information and ego status appears to yield results comparable to those achieved by employing ego status alone. Therefore, in models that integrate both ego vehicle status and perception information, a pertinent question arises: \textit{What specific role does the perception information, acquired from camera inputs, play within the final planning module?}

\begin{table*}[htb!]
\begin{center}
\resizebox{\textwidth}{!}{
\setlength{\tabcolsep}{9pt}
\begin{tabular}{l|cc|cccc|cccc|cccc}
\toprule

\multirow{2}{*}{Method} &
\multicolumn{2}{c|}{Ego Status} &
\multicolumn{4}{c|}{L2 (m) $\downarrow$} & 
\multicolumn{4}{c|}{Collision (\%) $\downarrow$} &
\multicolumn{4}{c|}{CCR (\%) $\downarrow$}  \\
 &in BEV&in Planer& 1s & 2s & 3s &\cellcolor{gray!30}Avg. & 1s & 2s & 3s& 
\cellcolor{gray!30}Avg. & 1s & 2s & 3s &\cellcolor{gray!30}Avg.\\
\midrule

% \midrule
 BEV-Planner &\xmark &\xmark & 0.30 & 0.52&0.83 &0.55 & 0.10 & 0.37 & 1.30 &0.59&0.78&3.79& 8.22&4.26\\
 BEV-Planner (init*) &\xmark&\xmark & 0.26 & 0.49 & 0.81 & 0.52 & 0.03 & 0.20 & 1.00 & 0.42& 0.59 & 3.18&7.36 & 3.71\\
 BEV-Planner+Map &\xmark &\xmark & 0.53 & 0.94&1.40 &0.96 & 0.12 & 0.37 & 2.19 & 0.89&0.68&2.38& 4.73&2.60\\
\bottomrule
\end{tabular}}
\end{center}

\vspace{-10pt}
\caption{While adding map perception task into the BEV-Planner method, we can observe that the model obtains worse L2 distance and collision rate performance, but achieves better CCR. We use a pretrained map perception checkpoint as the initialization of the BEV-Planner (init*)  and BEV-Planner+Map model.}
\label{tab:baseline_app3}
\vspace{-10pt}
 \end{table*}

\begin{table*}[htb!]
\begin{center}
\resizebox{\textwidth}{!}{
\setlength{\tabcolsep}{9pt}
\begin{tabular}{l|cc|cccc|cccc|cccc}
\toprule

\multirow{2}{*}{Method} &
\multicolumn{2}{c|}{Ego Status} &
\multicolumn{4}{c|}{L2 (m) $\downarrow$} & 
\multicolumn{4}{c|}{L2-ST (m) $\downarrow$} &
\multicolumn{4}{c|}{L2-LR (m) $\downarrow$}  \\
 &in BEV&in Planer& 1s & 2s & 3s &\cellcolor{gray!30}Avg. & 1s & 2s & 3s& 
\cellcolor{gray!30}Avg. & 1s & 2s & 3s &\cellcolor{gray!30}Avg.\\
\midrule

% \midrule
 BEV-Planner &\xmark &\xmark & 0.30 & 0.52&0.83 &0.55 & 0.27 & 0.47 & 0.78 &0.48 &0.43&0.78& 1.23&0.81\\
 BEV-Planner+Map &\xmark &\xmark & 0.53 & 0.94&1.40 &0.96 & 0.54 & 0.95 & 1.42 &0.97 &0.52&0.87& 1.28&0.89\\
 % \midrule
\bottomrule
\end{tabular}}
\end{center}

\vspace{-5pt}
\caption{L2-ST is the L2 distance with going straight driving commands. L2-LR is the L2 distance with turning left/right commands. }
\label{tab:baseline_app4}
\vspace{-10pt}
 \end{table*}

\begin{table*}[htb!]
\begin{center}
\resizebox{\textwidth}{!}{
\setlength{\tabcolsep}{9pt}
\begin{tabular}{l|cc|cccc|cccc|cccc}
\toprule

\multirow{2}{*}{Method} &
\multicolumn{2}{c|}{Ego Status} &
\multicolumn{4}{c|}{Collision (m) $\downarrow$} & 
\multicolumn{4}{c|}{Collision-ST (m) $\downarrow$} &
\multicolumn{4}{c|}{Collision-LR (m) $\downarrow$}  \\
 &in BEV&in Planer& 1s & 2s & 3s &\cellcolor{gray!30}Avg. & 1s & 2s & 3s& 
\cellcolor{gray!30}Avg. & 1s & 2s & 3s &\cellcolor{gray!30}Avg.\\
\midrule

% \midrule
 BEV-Planner &\xmark &\xmark & 0.10 & 0.37&1.30 &0.59 & 0.07 & 0.20 & 0.92 & 0.40&0.15&1.76& 4.84&2.25\\
 BEV-Planner+Map &\xmark &\xmark & 0.12 & 0.37&2.19 &0.89 & 0.14 & 0.38 & 2.20 &0.91 &0.00&0.29& 2.05&0.78\\
 % \midrule
\bottomrule
\end{tabular}}
\end{center}
\vspace{-5pt}
\caption{Collision-ST is the collision rate with going straight driving commands. Collision-LR is the collision rate with turning left/right commands.}
\label{tab:baseline_app5}
\vspace{-10pt}
 \end{table*}

\paragraph{Ego Status \vs~ Perceptual Information}
Undoubtedly, perceptual information constitutes the indispensable foundation of all autonomous driving systems, with ego status additionally offering crucial data such as the vehicle's velocity and acceleration to aid the system's decision-making process. Incorporating both perceptual information and ego status for the ultimate planning should indeed be a judicious strategy within an end-to-end autonomous driving system. However, as shown in \cref{tab:sota-plan}, relying solely on ego status can yield planning results that are on par with or even superior to those methods that utilize both ego status and perception modules on previous L2 or collation rate metrics. To ascertain the roles that perceptual information and ego status play in the final planning process, we introduced varying degrees of perturbation to the images and the ego status, as shown in the \cref{tab:noise}. We use the official VAD model (which leverages ego status in the planner module) as the base model.
It is observable that when disturbances are added to the images, the results of planning marginally decrease and may even exhibit improvement, while perceptual performance significantly deteriorates. 
Surprisingly, even when blank images are used as input, leading to the complete breakdown of the perception module, the model's planning capabilities remain largely unaffected.  The corresponding visualization results are as illustrated in the ~\cref{fig:vad_noise}. In contrast to the model's remarkable robustness to variations in image inputs, it exhibits considerable sensitivity to ego status. Upon altering the velocity of the ego car, we can observe that the planning results of the model are significantly impacted, as shown in \cref{fig:velo_noise}. Setting the ego car's speed to 100 m/s results in the model generating wildly impractical planning trajectories. We posit that an autonomous driving system displaying such heightened sensitivity to ego status information harbors considerable safety risks. 
Furthermore, with planning results being predominantly dictated by ego status, 
the functions of other modules in the model cannot be reflected. For example, while comparing VAD(ID-6) and BEV-Planner++ (ID-12), we can observe that they obtain basically similar results in terms of L2 and collision rate. Is it justifiable to assert that our BEV-Planner++ design, characterized by its simplicity and effectiveness, can attain comparable outcomes to other more intricate methodologies, even in the absence of utilizing perception data?
In fact, as the performance of the final planning module is predominantly influenced by the ego vehicle status, the design of other components does not significantly affect the planning results. Consequently, we argue that methods utilizing ego status are not directly comparable and conclusions should not be drawn from such comparisons.

\noindent\paragraph{\textit{How about not using ego status?}}
Given that the ego vehicle status exerts a dominant influence on the planning results, it prompts an important inquiry: Is it feasible and beneficial to exclude ego status in open-loop end-to-end research?

\paragraph{Neglected Ego Status in Perception Stage.} In fact, existing methods~\cite{jiang2023vad,ye2023fusionad} ignore the impact of using ego status on planning in BEV Encoder. More details are in the Appendix.

\paragraph{Without Ego Status, the Simpler, the Better?}
People might wonder why our BEV-Planner, without using additional perception tasks (including Depth, HD map, Tracking, \etc) and ego status, achieves better results in L2 distance and collision rate than other methods (ID-1 and 4). Since our BEV-Planner performs poorly in terms of CCR, what would happen if we added map perception tasks to our baseline?
To address these questions, we designed a ``BEV-Planner+Map" model by introducing a map perception task into our pipeline, mainly following the designs of UniAD.
As shown in \cref{tab:baseline_app3}, when map perception is introduced, the model exhibits poorer results in terms of L2 distance and collision rate metrics. The only aspect that aligns with our expectations is that the introduction of map perception significantly reduces the CCR. 
Through a comparison of BEV-Planner with BEV-Planner (init*), we observe that the use of map-pretrained weights can enhance performance. This finding implies that the decrease in L2 and Collision rate observed with the integration of MapFormer in ``BEV-Planner+Map" is not due to the pretrained weights. We posit that in most straight-driving scenarios, the addition of lane information may not yield markedly effective information and could indeed introduce some degree of interference.
To verify our hypothesis, we evaluated the performance of these methods under varying driving commands. As shown in \cref{tab:baseline_app4} and \cref{tab:baseline_app5}, adding map information significantly increases the L2 distance error and collision rate with going straight commands. In contrast, for turning scenarios, the incorporation of map information effectively reduces the collision rate.
Based on the above observations, we can tentatively draw the following conclusions:
\begin{itemize}
    \item In simple straightforward driving scenarios, the addition of perceptual information does not appear to enhance the model's performance with respect to L2 distance and collision rate. Conversely, the implementation of more intricate multi-task learning paradigms may, in fact, lead to a decrease in the model's overall efficacy.
    \item  In more complex scenarios, such as turns, incorporating perceptual information can be beneficial for planning purposes. However, given the relatively small proportion (13\%) of turning scenes in the existing evaluation datasets, the introduction of perceptual information tends to adversely affect the average performance metrics (L2 distance and collision rate) in the final analysis.
    \item It is imperative to develop a more robust and representative evaluation dataset. The metrics derived from the current evaluation dataset are not entirely persuasive and fail to accurately reflect the true capabilities of the model.

\end{itemize}

\begin{table}[t]
\centering

\resizebox{\columnwidth}{!}{\begin{tabular}{l|c|c|c|c} 
%\begin{tabular}{ p{10mm}<{\centering}| p{5mm}<{\centering} | p{12mm}<{\centering}  p{10mm}<{\centering}  p{10mm}<{\centering} }
\toprule
{Method} & \
{P.P.}  & Avg. L2(m)  & Avg. Colli.(\%) & Avg. CCR(\%)
\\
\midrule
UniAD  &\cmark 
& 0.77 & \textbf{0.51} & 7.83\\
UniAD & \xmark 
& \textbf{0.66} & 0.62 & \textbf{1.72}\\
\bottomrule
\end{tabular}}
\caption{ P.P. indicates the post-processing optimization module of UniAD.
We use the officially released weights of UniAD for this ablation study.
By default, UniAD incorporates a post-processing step to refine the predicted trajectory from the end-to-end model, aiming to mitigate collisions with other agents. Nonetheless, this approach is limited by this singular optimization objective, which overlooks additional safety-critical factors in autonomous driving, such as lane adherence. Our new metrics, the CCR reveal that UniAD's post-processing substantially increases the risk of the ego vehicle running off the road.}
\vspace{-1mm}
\label{tab:abl_uniad}
\vspace{-10pt}
\end{table}

\paragraph{New metrics will bring new conclusions.}
The preceding methodology primarily centered around the L2 distance and collision rate metrics. Our discussion thus far has been largely concentrated on these two metrics. What we want to emphasize is that these two metrics, L2 distance and collision rate, only reflect a partial aspect of a model's planning capabilities. It is not advisable to assess the quality of a model based exclusively on these two metrics. In this paper, we introduce a new metric to evaluate the model's comprehension and adherence to the map: Curb Collision Rate (CCR). As shown in \cref{tab:sota-plan}, we can observe that the GoStright strategy frequently intersects with road boundaries, which is in line with our expectations. In terms of this new metric, Ego-MLP performs worse than UniAD and VAD. Our method BEV-Planner, performs the worst on this metric because it does not use any map information.
This suggests that relying on past metrics to judge the superiority of different open-loop methods is biased.

Based on our proposed new metrics, we also found that the existing collision rate metric can be manipulated with post-processing.  More specifically, within UniAD~\cite{hu2023uniad}, a non-linear optimization module is employed to refine the trajectory predicted by the end-to-end model, ensuring that the anticipated path steers clear of the occupancy grid, thereby aiming to prevent collisions. However, this optimization, while significantly reducing the collision rates with other agents, inadvertently introduces additional safety risks. The absence of adequate constraints in its optimization process, such as the integration of map priors, markedly increases the risk of the optimized trajectory encroaching upon road boundaries, as shown in \cref{tab:abl_uniad}. In this paper, we report the results of UniAD without its post-processing by default.

What we wish to underscore is that the primary intention behind proposing CCR is to illuminate the inadequacies within the existing evaluation systems. However, even with the incorporation of CCR, open-loop evaluation systems still encounter numerous challenges. We assert that the evaluation of open-loop autonomous driving systems necessitates a more diverse and stringent evaluation framework.
This would enable a more accurate reflection of these systems' capabilities and limitations.

\paragraph{What the baseline learned in its BEV?}
As shown in~\cref{fig:convg}, with the influence of ego status, the models converge rapidly. Considering the challenge of generating valuable BEV features from visual images and comparing the convergence curve of BEV-Planer that does not use ego status, this further demonstrates that ego status information dominates the learning process. Since our baselines are solely supervised by ego trajectory, we are wondering what the model is learning from the images. As shown in \cref{fig:bev_scenes}, we observed a distinct phenomenon: in BEV-Planner++, the activation range of the feature map predominantly encompasses the immediate vicinity around the ego vehicle, frequently manifesting behind the vehicle itself. This pattern marks a significant deviation from the BEV-Planner's BEV features, which typically concentrate on the area ahead of the vehicle. We speculate that this is due to the introduction of ego status information, which negates the model's need to extract information from BEV features. Hence, the BEV-Planner++ method 
has almost not learned any effective information.
\vspace{-10pt}

\section{Conclusion}
In this paper, we present an in-depth analysis of the shortcomings inherent in current open-loop, end-to-end autonomous driving methods.
Our objective is to contribute findings that will foster the progressive development of end-to-end autonomous driving.

\vspace{2mm}
\noindent Our conclusions are summarized as follows:
\vspace{2mm}
\begin{itemize}
    \item The planning performance of existing open-loop autonomous driving models based on nuScenes is highly affected by ego status (velocity, acceleration, yaw angle). With ego status involved, the model's final predicted trajectories are basically dominated by it, resulting in a diminished use of sensory information.
    \item Existing planning metrics fall short of fully capturing the true performance of models. The evaluation results of the model may vary significantly across different metrics. We advocate for the adoption of more diverse and comprehensive metrics to prevent models from achieving local optimality on specific metrics, which may lead to the neglect of other safety hazards.
    \item Compared to pushing the state-of-the-art performance on the existing nuScenes dataset, we assert that the development of more appropriate datasets and metrics represents a more critical and urgent challenge to tackle.
    
    \end{itemize}
\vspace{-15pt}
\paragraph{Limitation}

There are trade-offs between different planning metrics. Designing an integrated evaluation system for open-loop evaluation presents a significant challenge. Although our baseline method excels in terms of L2 distance and collision rate, its performance is not exceptional in the CCR, primarily because our approach does not utilize any perception annotations, such as HD maps.

\section*{Acknowledgments}

Thanks to Bencheng Liao and Shaoyu Chen for providing helpful information and the model weights of VAD. Tong Lu and Zhiqi Li are supported by the National Natural Science
Foundation of China (Grant No. 62372223) and China Mobile Zijin Innovation Insititute (No. NR2310J7M). Zhiqi Li is also supported by the NVIDIA Graduate Fellowship Program.


{
    \small
    \bibliographystyle{ieeenat_fullname}
    \bibliography{main}
}
\appendix
\setcounter{figure}{0}
\setcounter{table}{0}
\maketitlesupplementary

\section{Implementation Details}

\paragraph{ST-P3 uses partially incorrect training and evaluation data.}
For the common practice, the future GT planning trajectory is generated from the ego locations of the samples in the subsequent 3 seconds. However, since one nuScenes clip is usually a 20s video, which means that the samples at the tail of the video (within 17s-20s) cannot produce a complete future trajectory, normal methods~\cite{jiang2023vad,hu2023uniad} will perform special processing on these special samples by using masks, but ST-P3~\cite{hu2022stp3} did not do this. ST-P3 mistakenly used samples from other scenes while generating GT of these tail samples, so errors occurred during training and testing. Related issue: \url{https://github.com/OpenDriveLab/ST-P3/issues/24}.

\paragraph{Ego Status Usage Details}
For UniAD (ID-1) and VAD-Base (ID-4), in order to exclude ego status from the Bird's Eye View (BEV) generation phase, we set the use\_can\_bus flag to False. Conversely, for UniAD (ID-3), to incorporate ego status into its planner, we adhered to the methodology used in VAD, which involves concatenating the ego status vector with the query features.

\section{Metrics Details.}

\begin{figure}[t]
	\centering
	\includegraphics[width=1\columnwidth]{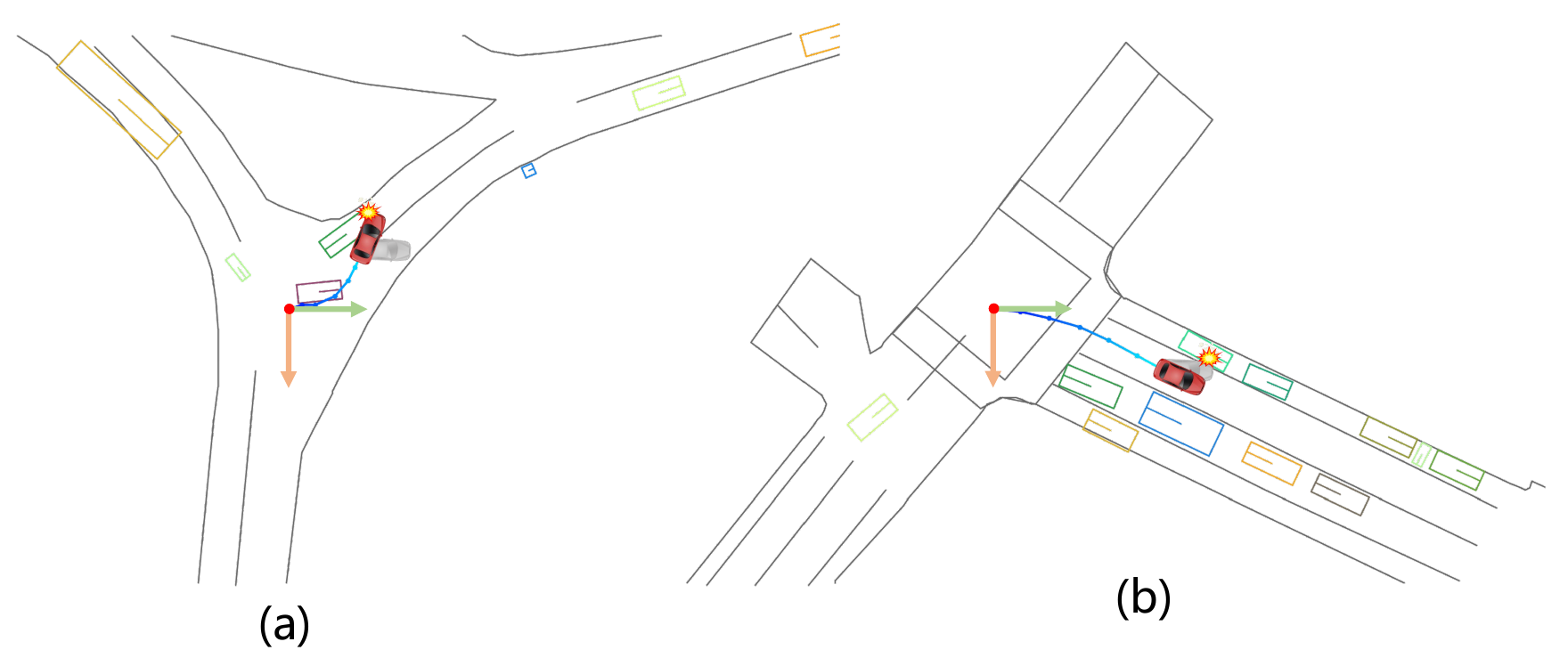}
   % \vskip -0.1in
   %\vspace{-3mm}

   \caption{
   Current methods~\cite{hu2023uniad,jiang2023vad, hu2022stp3} neglect to consider yaw angle variations of the ego vehicle, consistently preserving a 0 yaw angle (depicted by the gray vehicle), thereby resulting in an increased incidence of false negatives (a) and false positives in (b) collision detection. In this paper,  we improve collision detection accuracy by estimating the vehicle's yaw angle from variations in its trajectory (depicted by the red vehicle).}
   \label{fig:colli}
  \vspace{-8pt}
   \end{figure}

\begin{figure}[t]
	\centering
	\includegraphics[width=1\columnwidth]{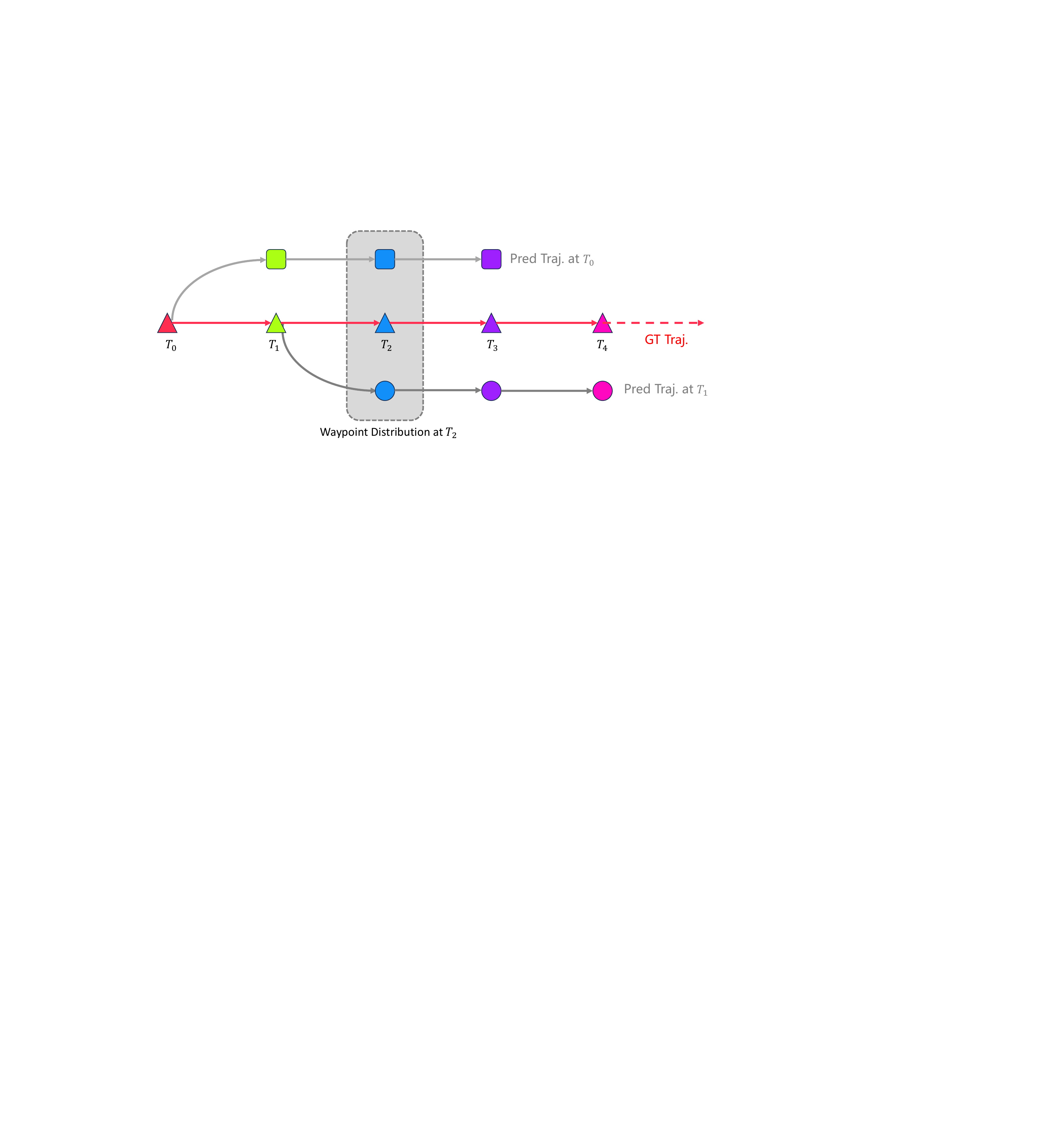}
   % \vskip -0.1in
   %\vspace{-3mm}
   \caption{
In open-loop autonomous driving approaches, the future trajectory is forecasted from the starting location of the ego vehicle. Within the imitation learning paradigm, the predicted trajectory ideally should closely align with the actual ground truth trajectory. Furthermore, trajectories forecasted at successive time steps should maintain consistency, thereby guaranteeing the continuity and smoothness of the driving strategy. Consequently, the predicted trajectories depicted in red boxes of \cref{fig:vad_noise} not only deviate from the ground truth trajectory but also demonstrate significant divergence at various timestamps.}
   \label{fig:traj_tree}
   \vspace{-20pt}
   \end{figure}
\paragraph{Collision Rate.} While current methods tend to evaluate the collision rates of planned trajectories~\cite{hu2021safe, khurana2022differentiable, hu2022stp3, hu2023uniad,jiang2023vad, ye2023fusionad}, there are issues in both the definition and implementation of this metric in existing approaches. First of all, in open-loop end-to-end autonomous driving, other agents do not provoke a response from the ego car. Instead, they strictly adhere to their predetermined trajectories. Consequently, this leads to a bias in the calculation of collision rates. The second issue arises from the fact that the planning predictions generated by current methods consist solely of a series of trajectory points. As a consequence, in the final collision calculation, the yaw angle of the ego car is not taken into account. Instead, it is assumed to remain unchanged. This assumption leads to erroneous results, particularly in turning scenarios, as shown in \cref{fig:colli}. 

There are also problems in the current implementation.
The current definition of the collision rate of each single sample is:
\vspace{-2pt}
\begin{equation}
\label{ori_col_rate}
CR(t) = \frac{\sum_{i=0}^{N}\mathbb{I}_{i}}{N}, N = t/0.5,
\end{equation}
\vspace{-2pt}
$N$ represents the number of steps at intervals of $t$ seconds, and $\mathbb{I}_i$ denotes whether the ego car at step $i$ will intersect with other agents. 
In this paper, we modify the definition of collision  to
\begin{equation}
\label{our_col_rate}
CR(t) = {(\sum_{i=0}^{N}\mathbb{I}_{i}})>0, N = t/0.5.
\end{equation}

For previous implementation, they assumed that collisions at each moment were mutually independent, which does not align with real-world scenarios. Our modified version yields values that more precisely indicate the collision rate occurring along the predicted trajectory.

\begin{table}[htb!]
\begin{center}
\resizebox{0.5\textwidth}{!}{
\setlength{\tabcolsep}{5pt}
\begin{tabular}{l| cccc|cccc}
\toprule
\multirow{2}{*}{Method} &
\multicolumn{4}{c|}{L2 (m) $\downarrow$} & 
\multicolumn{4}{c|}{($\sigma_{wd}$) $\downarrow$} \\
 & 1s & 2s & 3s & \cellcolor{gray!30}Avg. & 1s & 2s & 3s & \cellcolor{gray!30}Avg.\\
\midrule
Baseline &{0.30} & {0.52} & {0.85} & {0.56}  & 0.03 & 0.19 & 0.70 & 0.31\\
\bottomrule
\end{tabular}
}
\end{center}
\vspace{-10pt}
\caption{The smoothness $\sigma_{wd}$ of predicted trajectories.}

\label{tab:smoothness}
\vspace{-10pt}
\end{table}
\paragraph{Trajectory Smoothness}
We also assessed the stability of the model's predicted trajectories. Given that the model predicts the trajectory for the next three seconds at each moment, it means that for every absolute moment in time $t$, the model predicted multiple waypoints at time $t$ from various preceding times. 
We see these different waypoints as a distribution. In non-extreme conditions, this distribution should be as concentrated as possible to ensure smoothness in the driving process, as shown in~\cref{fig:traj_tree}. To quantitatively analyze this distribution, we calculated the squared deviation distance of these distribution points, as shown in ~\cref{tab:smoothness}. We found that this smoothness metric does not convey more information than the L2 metric, and we believe that this requires more exploration to verify the rationality of an metric.

\paragraph{Valid Samples}
We discussed above that for the tail samples without complete GT trajectories. The normal method will use the mask for special identification. During evaluation, the previous methods have different processing methods. One is that if a sample does not have a complete GT future trajectory, it will not be considered during evaluation. The second strategy only considers the valid part of the GT future trajectory if the length of the GT trajectory is less than 3s. In this paper, we follow the first strategy. For 6019 samples of nuScenes val split, the number of final valid samples is 5119 (85\% of all samples). The reason why we didn't reproduce the correct version of ST-P3 is that the definition of valid samples of ST-P3 is different from others.  Valid samples of ST-P3 must use sufficient historical data, so ST-P3 does not predict trajectories for the first few samples of each clip. Even if we reproduce the correct ST-P3, we cannot compare it with other methods for a fair comparison.

\begin{figure}[htb]
	\centering
	\includegraphics[width=\columnwidth]{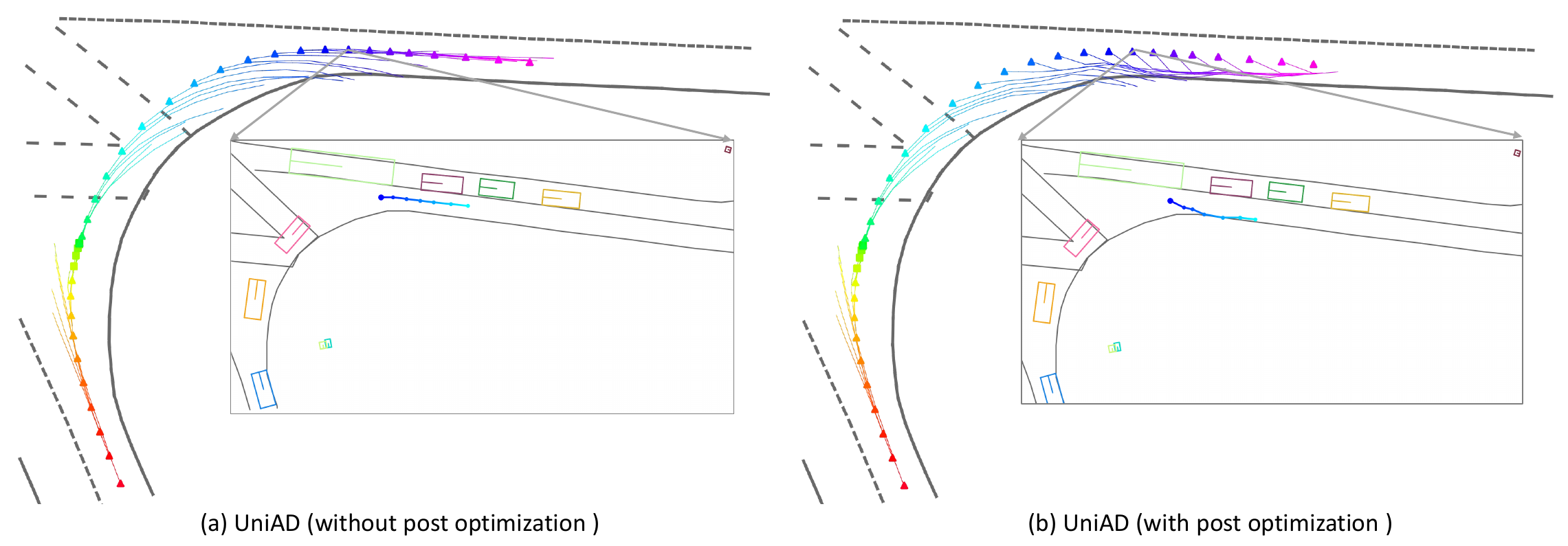}
   % \vskip -0.1in
   \caption{In order to avoid collisions as much as possible, post-optimization is introduced in UniAD~\cite{hu2023uniad} to keep the predicted trajectory away from other vehicles. However, during actual traffic driving, other factors need to be considered, such as road conditions. As shown in figure (b), UniAD rushed to the road boundary in order to avoid the possible danger caused by the opposite lane and actually caused another accident.
   }
  %  \vspace{-0.20in}
\label{fig:uniad_pp}
\end{figure}

\section{Neglected Ego Status in Perception Stage.}\label{ego_status_in_perception_stage}
\begin{figure}[t]
	\centering
	\includegraphics[width=0.8\columnwidth]{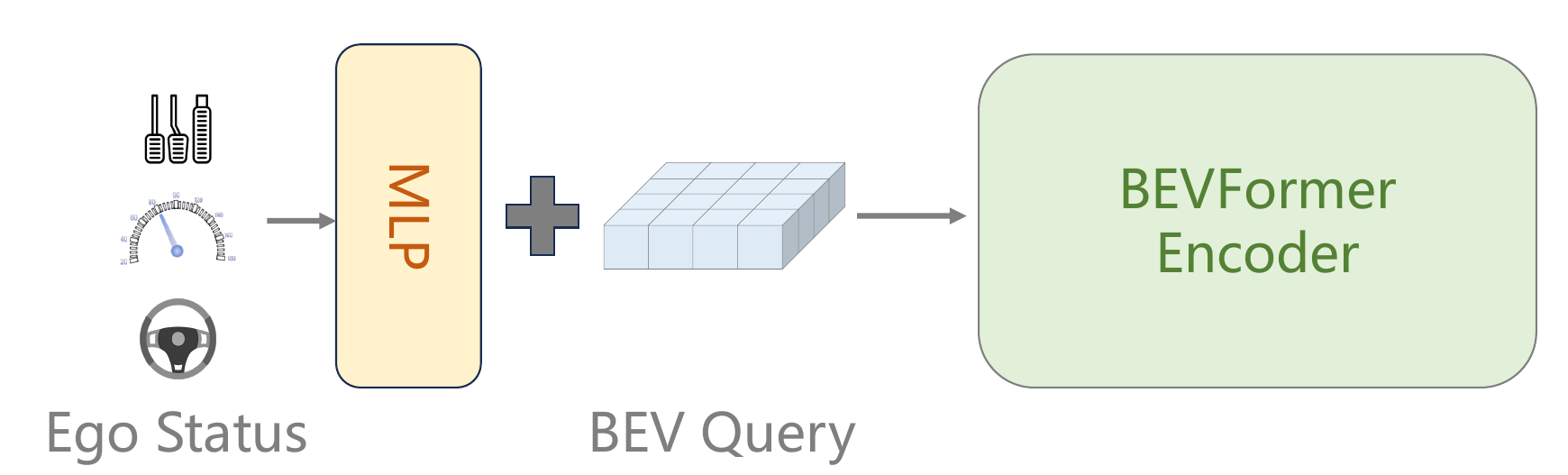}
   % \vskip -0.1in
   %\vspace{-3mm}
   \caption{BEVFormer incorporates ego status information during the initialization of BEV queries, a nuance not addressed by current end-to-end autonomous driving approaches~\cite{jiang2023vad,hu2023uniad,ye2023fusionad}.}
  \vspace{-10pt}
\label{fig:bevformer}
\end{figure}

In fact, a crucial question is whether the method really completely eliminates the influence of ego status.
The pipeline of the existing open-loop end-to-end autonomous driving methods~\cite{hu2023uniad, jiang2023vad, ye2023fusionad, hu2022stp3} basically follows the  ~\cref{fig:e2e_pipeline} (b). 
Given that ego status exerts a substantial influence on the planning results, these methods actually have clear explanations on whether to introduce ego status in the planner. 
However, methods~\cite{hu2023uniad, jiang2023vad} ignored the impact of introducing ego status in the early perception stage on the planning results.
In detail, both UniAD~\cite{hu2023uniad} and VAD~\cite{jiang2023vad} utilize BEVFormer~\cite{li2022bevformer} as their BEV generation module. For BEVFormer, it involves projecting the ego status onto the hidden features and incorporating it into the BEV query, as shown in \cref{fig:bevformer}.
This trick exerts a marginal effect on perception performance, as shown in \cref{tab:abl_bevformer}. However, when BEVFormer is integrated into an end-to-end pipeline, the introduction of ego status at this initial stage can wield a substantial influence on the ultimate planning performance.  As shown in \cref{tab:sota-plan}, upon the removal of the ego status input during the BEV stage, the planning performance of both VAD and UniAD exhibits a marked decline. 
It is important to clarify that our position is not opposed to the use of ego status; rather, we argue that within the context of current datasets and evaluation metrics, the integration of ego status can significantly impact, and even determine, the planning results. Unfortunately, the incorporation of ego status within the perception module is often overlooked in the existing end-to-end autonomous driving methods. Therefore, it is essential in comparative analyses of different methodologies to carefully examine the role and impact of ego status to ensure fairness and consistency in the evaluations.
\begin{table}[htb]
\centering

\vspace{-1mm}
\setlength{\tabcolsep}{15pt}
\resizebox{\columnwidth}{!}{\begin{tabular}{l|c|c|c
} 
% \begin{tabular}{ p{25mm}<{\centering}| p{10mm}<{\centering} | p{12mm}<{\centering}  p{10mm}<{\centering}  p{10mm}<{\centering} p{10mm}<{\centering} p{10mm}<{\centering} p{12mm}<{\centering} | p{10mm}<{\centering}}
\toprule
{Methods} & \
{Ego Status}  & {mAP}$\uparrow$  &{NDS}$\uparrow$  
\\
\midrule
BEVFormer  &\cmark 
&41.6 & 51.7\\
BEVFormer & \xmark 
& 41.3 & 51.5\\
\bottomrule
\end{tabular}}
\caption{
The integration of ego status  within BEVFormer exerts only a marginal effect on the perception performance.}
\vspace{-5pt}
\label{tab:abl_bevformer}\end{table}
\section{Post Optimization of UniAD.}

As demonstrated in \cref{fig:uniad_pp}, we observe that while UniAD utilizes collision optimization, the resulting optimized trajectory tends to intersect with the road boundary at a higher rate. This occurs because the collision optimizer overlooks map priors. In its effort to avoid collisions, the optimizer disregards other factors that could pose safety risks. However, if the optimizer were to consider all relevant factors, it would more closely resemble traditional Planning and Navigation Control (PNC) systems, contradicting the fundamental motivation of end-to-end autonomous driving.

\section{Dropping Cameras}
\begin{figure*}[htb!]
	\centering
	\includegraphics[width=\textwidth]{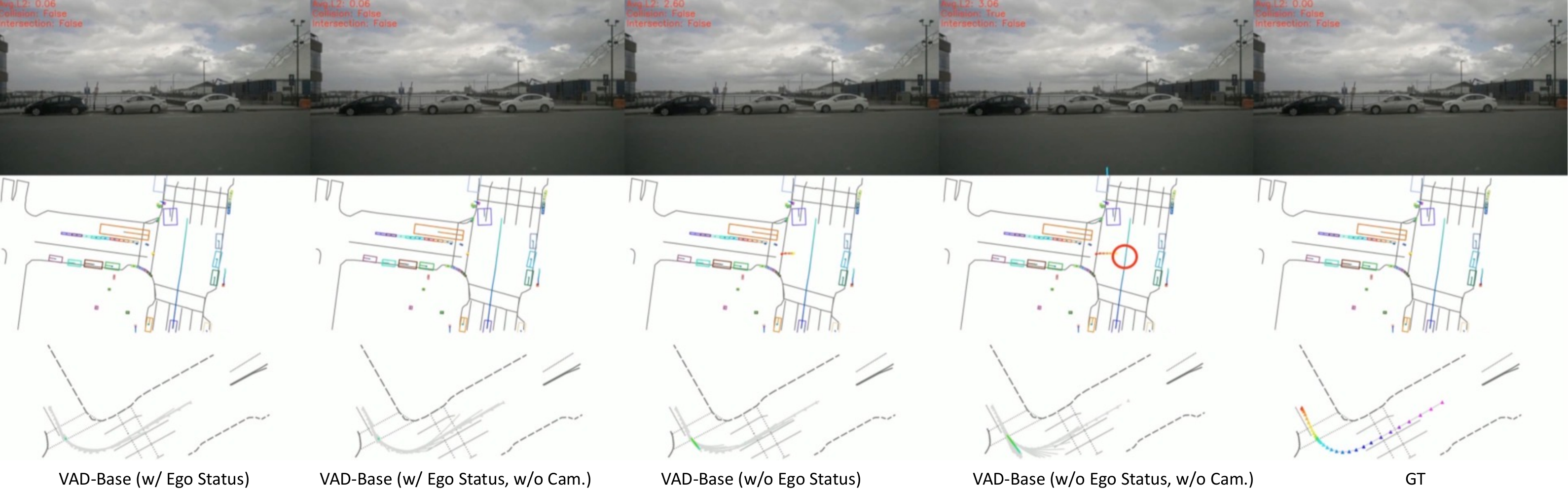}
   % \vskip -0.1in
   \caption{
   When the model uses ego status as an input, removing the camera does not significantly impact its performance. However, without ego status, omitting camera inputs makes the model more prone to erroneous planning. Red circles indicate potential collisions.}
\vspace{-10pt}
\label{fig:drop_pp}
\end{figure*}

Referencing Table 2 in our main paper, it is observed that when VAD incorporates ego status as an input, the removal of camera input does not markedly impair its performance. A parallel experiment was conducted with VAD~\cite{jiang2023vad} devoid of ego status. We also provide visualization results in \cref{fig:drop_pp}. As delineated in \cref{tab:noise_app}, excluding camera inputs in VAD without ego status leads to a significant decline in performance, particularly regarding L2 distance and collision rate metrics. Intriguingly, this decrease was not mirrored in the Intersection rate with road boundary metric.

\begin{table*}[htb!]
\begin{center}
\resizebox{\textwidth}{!}{
\setlength{\tabcolsep}{5pt}
\begin{tabular}{l|c|c| cccc|cccc|cccc|cc}
\toprule
\multirow{2}{*}{Method} &
{Img } &
\multirow{2}{*}{Ego Status} &
\multicolumn{4}{c|}{L2 (m) $\downarrow$} & 
\multicolumn{4}{c|}{Collision (\%) $\downarrow$} &
\multicolumn{4}{c|}{Intersetion (\%) $\downarrow$} &
Det. & Map\\
 &Corruption&& 1s & 2s & 3s & \cellcolor{gray!30}Avg. & 1s & 2s & 3s & \cellcolor{gray!30}Avg.& 1s & 2s & 3s & \cellcolor{gray!30}Avg. &(NDS) &(mAP) \\
\midrule
VAD-Base & - & \cmark& \textbf{0.17} & \textbf{0.34} & \textbf{0.60} & \textbf{0.37} & 0.04 & \textbf{0.27} & \textbf{0.67}&\textbf{0.33}&\textbf{0.21}&
\textbf{2.13}&\textbf{5.06} & \textbf{2.47}& \textbf{45.5} &  47.0\\
VAD-Base &  Blank & \cmark& 0.19 & 0.41 & 0.77 & 0.46 & \textbf{0.00} & {0.40} & {1.21}&0.54 &0.35&3.05&7.73&3.71& 0.0 &0.0  \\% 

\midrule
VAD-Base & - & \xmark& {0.69} & {1.22} & {1.83} & {0.06} & 0.68 & {2.52} & {0.84}&0.37&1.02&3.44&7.00 & 3.82& 45.1 &  \textbf{53.7}\\
VAD-Base &  Blank & \xmark& 2.59 & 4.32 & 6.09 & 4.33 & {2.29} & {7.89} & {12.7}&7.63 &1.07&3.73&6.64&3.81 &0.0 &0.0  \\% 

\bottomrule

\end{tabular}}
\end{center}
\vspace{-10pt}
\caption{
Omitting camera inputs in the VAD model, when it does not utilize ego status, results in a marked reduction in performance, as evidenced by the metrics for L2 distance and collision rate.}

\label{tab:noise_app}
\vspace{-10pt}
\end{table*}
  
\begin{table*}[htb!]
\begin{center}
\resizebox{\textwidth}{!}{
\setlength{\tabcolsep}{5pt}
\begin{tabular}{l|c|c| cccc|cccc|cccc|cc}
\toprule
\multirow{2}{*}{Method} &
{Img } &
\multirow{2}{*}{Ego Status} &
\multicolumn{4}{c|}{CCR $\downarrow$} & 
\multicolumn{4}{c|}{CCR-ST (\%) $\downarrow$} &
\multicolumn{4}{c|}{CCR-LR (\%) $\downarrow$} &
Det. & Map\\
 &Corruption&& 1s & 2s & 3s & \cellcolor{gray!30}Avg. & 1s & 2s & 3s & \cellcolor{gray!30}Avg.& 1s & 2s & 3s & \cellcolor{gray!30}Avg. &(NDS) &(mAP) \\
\midrule
VAD-Base & - & \xmark& {1.02} & {3.44} & {7.00} & {3.82} & 2.49 & {8.50} & {16.4}&9.13 &0.95&2.70&5.50 & 3.05& 45.1 &  {53.7}\\
VAD-Base &  Blank & \xmark& 1.07 & 3.73 & 6.64 & 3.81 & {2.63} & {18.2} & {32.1}&17.6 &0.83&1.51&2.72&1.69 &0.0 &0.0  \\% 

\bottomrule

\end{tabular}}
\end{center}
\vspace{-10pt}
\caption{
CCR-ST is the CCR rate with going straight driving commands. CCR-LR is the CCR rate with turning left/right commands. 
}

\label{tab:noise_app2}
\vspace{-10pt}
\end{table*}

In fact, when the model operates without using ego status and with the camera input removed, it relies solely on driving commands to guide its future direction. In this scenario, the fact that the intersection rate with road boundaries does not increase is counter-intuitive. This counter-intuitive phenomenon has driven us to delve deeper into the evaluation process.
As shown in \cref{tab:noise_app2}, we compiled statistics on the intersection metric under different driving commands. The Intersection-LR metrics show that the model, when operating without camera input, significantly increases the probability of interacting with boundaries in turning scenarios.
This is also consistent with our observations from visualization.
The real reason lies in the fact that in straight-driving scenarios  (87\% of all evaluation samples), removing the camera input leads the model to adopt a relatively conservative straight-driving strategy, making it less likely to intersect with road boundaries (indicated by Intersection-ST). Since straight-driving scenarios constitute a large proportion of the \textit{val} split, this results in the model achieving better overall average results when operating without camera input.

\begin{figure*}[htb]
\centering
\includegraphics[width=0.9\textwidth]{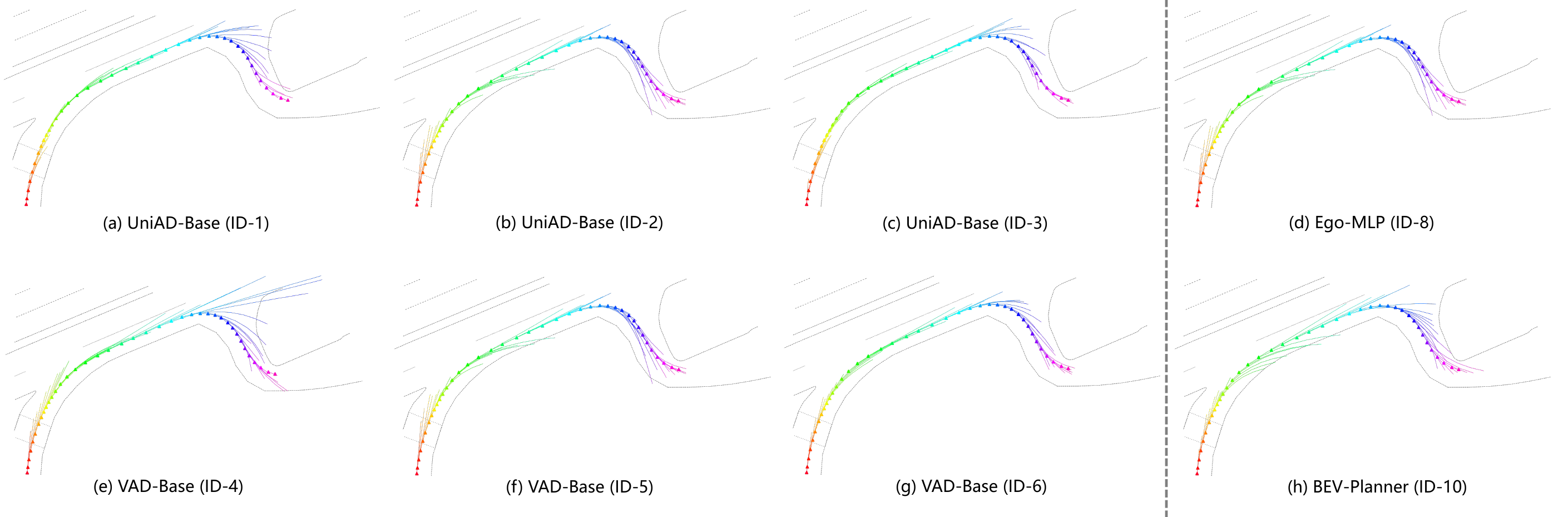}
t\caption{In scenarios that necessitate continuous turning, all methods predict suboptimal trajectories. }
\vspace{-10pt}
% \vspace{-0.20in}
\label{fig:turning}
\end{figure*}

\paragraph{Failure Cases}
Although the majority of scenarios in the nuScenes dataset are relatively straightforward, it does include certain challenging scenes, notably those involving continuous cornering. As shown in \cref{fig:turning}, we can observe that methods with various settings all yielded suboptimal predicted trajectories when navigating high-curvature bends.
For challenging scenarios like cornering, where the system must continuously make evolving decisions, evaluating open-loop autonomous driving systems poses a significant challenge. One limitation of open-loop methods is that they do not suffer from cumulative errors. 
In detail, in the case of an extremely erroneous trajectory predicted at a given timestep, the trajectory starting point for the next timestep is still based on the GT trajectory. The metric we utilize, CCR, is adept at identifying low-quality trajectories. However, an appropriate metric that can effectively highlight high-quality trajectories remains an intriguing direction for further exploration.

%\section{Rationale}
% \label{sec:rationale}
% % 
% Having the supplementary compiled together with the main paper means that:
% % 
% \begin{itemize}
% \item The supplementary can back-reference sections of the main paper, for example, we can refer to \cref{sec:intro};
% \item The main paper can forward reference sub-sections within the supplementary explicitly (e.g. referring to a particular experiment); 
% \item When submitted to arXiv, the supplementary will already included at the end of the paper.
% \end{itemize}
% % 
% To split the supplementary pages from the main paper, you can use \href{https://support.apple.com/en-ca/guide/preview/prvw11793/mac#:~:text=Delete%20a%20page%20from%20a,or%20choose%20Edit%20%3E%20Delete).}{Preview (on macOS)}, \href{https://www.adobe.com/acrobat/how-to/delete-pages-from-pdf.html#:~:text=Choose%20%E2%80%9CTools%E2%80%9D%20%3E%20%E2%80%9COrganize,or%20pages%20from%20the%20file.}{Adobe Acrobat} (on all OSs), as well as \href{https://superuser.com/questions/517986/is-it-possible-to-delete-some-pages-of-a-pdf-document}{command line tools}.

%\newpage
\end{document}